\documentclass[letterpaper, 10 pt, conference]{ieeeconf}  %

\IEEEoverridecommandlockouts                              %

\usepackage{xcolor}
\usepackage[nolist,nohyperlinks]{acronym}
\usepackage{amssymb}
\usepackage{amsmath}
\usepackage{graphicx}
\usepackage{gensymb}
\usepackage{array}
\usepackage{colortbl}

\newcommand\method{CFTrack}

\acrodef{mot}[MOT]{Multi Object Tracking}
\acrodef{dnn}[DNN]{Deep Neural Networks}
\acrodef{cnn}[CNN]{Convolutional Neural Networks}
\acrodef{mota}[MOTA]{Multi-Object Tracking Accuracy}
\acrodef{dla}[DLA]{Deep Layer Aggregation}

\title{\LARGE \bf
CFTrack: Center-based Radar and Camera Fusion for 3D Multi-Object Tracking
}

\author{Ramin Nabati, Landon Harris, and Hairong Qi%
\thanks{The authors %
are with the Department of Electrical Engineering and Computer Science, University of Tennessee, Knoxville, TN, USA. Email: 
        {\tt\small rnabati@utk.edu, lharri73@vols.utk.edu, hqi@utk.edu}}%
}

\begin{document}
\maketitle
\thispagestyle{empty}
\pagestyle{empty}

\begin{abstract}
3D multi-object tracking is a crucial component in the perception system of autonomous
driving vehicles. Tracking all dynamic objects around the vehicle is essential 
for tasks such as obstacle avoidance and path planning. Autonomous vehicles are usually
equipped with different sensor modalities to improve accuracy and reliability. While
sensor fusion has been widely used in object detection networks in recent years, most 
existing multi-object tracking algorithms either rely on a single input modality, or 
do not fully exploit the information provided by multiple sensing modalities. In 
this work, we propose an end-to-end network for joint object detection and tracking 
based on radar and camera sensor fusion. Our proposed method uses a center-based
radar-camera fusion algorithm for object detection and utilizes a greedy
algorithm for object association. The proposed greedy algorithm uses the depth, 
velocity and 2D displacement of the detected objects to associate them through time. 
This makes our tracking algorithm very robust to occluded and overlapping objects, as
the depth and velocity information can help the network in distinguishing them.
We evaluate our method on the 
challenging nuScenes dataset, where it achieves 20.0 AMOTA and outperforms all vision-based 
3D tracking methods in the benchmark, as well as the baseline LiDAR-based method.
Our method is online with a runtime of 35ms per image, making it 
very suitable for autonomous driving applications.

\end{abstract}

\section{Introduction}
\ac{mot} is the task of analyzing videos to identify and track objects belonging
to certain categories, without any prior knowledge about the appearance or the 
number of targets \cite{ciaparrone2020deep}. Occlusions and interactions between 
objects with similar appearances are two main factors that make \ac{mot} a 
challenging task. Many algorithms have been developed 
in recent years to address these issues. The majority of these algorithms 
exploit the rich representational power of \ac{dnn} to extract complex semantic 
features from the input. Tracking-by-detection is a common approach used in these 
algorithms, where the tracking problem is solved by breaking it into two steps: 
(1) detecting objects in each image, (2) associating the detected objects over 
time. Recently, the \ac{cnn}-based object detection networks have been very successful
in improving the performance in this task. As a result, many of the 
\ac{mot} methods adopt an existing detection method and focus more on improving the
association step.

Object tracking is an important task in autonomous driving vehicles. Tracking of
dynamic objects surrounding the vehicle is essential for many of the tasks crucial
to autonomous navigation, such as path planning and obstacle avoidance \cite{rangesh2019no}. 
To increase reliability and accuracy, the perception system in an autonomous vehicle 
is usually equipped with multiple sensors with different sensing modalities such as 
cameras, radars and LiDARs. %
Incorporating the multi-modal sensory data 
into an object tracking framework for 
autonomous driving applications is not a trivial task. It requires an efficient, accurate 
and reliable fusion algorithm capable of utilizing the information embedded in different 
modalities in real time. Most multi-modal \ac{mot} methods 
use multiple sensing modalities in the detection stage, but only utilize features from
one sensing modality in the association step. In addition, many existing \ac{mot} methods rely only on camera images \cite{zhou2020tracking,zhu2018online} or LiDAR point clouds \cite{choi2013multi,song2015object} for detection and tracking. 

In recent years, radars have been widely used in vehicles for Advanced Driving
Assistance System (ADAS) applications such as collision avoidance
\cite{nabati2020centerfusion}. Radars are capable of detecting objects at much 
longer range compared to LiDAR and cameras, while being very robust to adverse 
weather conditions such as fog and snow. Additionally, radars provide accurate 
velocity information for every detected object. While objects' velocity information
might not be necessary for object detection, it is extremely useful for the 
object tracking task as it can be used for predicting objects' path and displacement.
Finally, compared to LiDARs, radar point clouds require less processing before 
they can be used as object detection results \cite{nabati2020centerfusion}. 
This is an important factor in real-time applications with limited processing 
power such as autonomous driving.

We propose an end-to-end \ac{mot} framework, utilizing radar and camera data to 
perform joint object detection and tracking. Our method is based on CenterFusion
\cite{nabati2020centerfusion}, a radar-camera sensor fusion algorithms for 3D object
detection in autonomous driving applications. CenterFusion achieves state-of-the-art
performance in 3D object detection using Radar and camera fusion and provides 
object velocity estimates that could be very helpful in the object tracking task.
Our proposed network takes as input the 
current image frame and radar detections in addition to the previous frame and detected
objects. The outputs are 3D object detection 
results and tracking IDs for the detected objects. Every detected 
object is also associated with an estimated absolute velocity in the global coordinate 
system. 

The object association step in our tracking framework is based on a simple greedy 
algorithm similar to CenterTrack \cite{zhou2020tracking}. While CenterTrack only 
uses the objects' 2D displacement in consecutive images to associate them, we propose 
a greedy algorithm based on a weighted cost function calculated from the object's 
estimated depth and velocity in addition to their 2D displacement. This significantly
improves the ability of the network to correctly associate occluded and overlapping 
objects, as the depth and velocity information provide valuable clues to distinguish 
these objects. Additionally, the proposed network uses the fused radar and image
features to predict the objects' displacement in consecutive frames, which makes these
predictions more accurate compared to just using the visual information. We thus refer to the proposed \ac{mot} framework as \method{}.

The main contributions of this paper are twofold. First, to the best knowledge of the authors,
this study is the first to propose a radar and camera sensor fusion framework for 
3D multi-object detection and tracking using a deep network trained end-to-end. 
Second, we propose a 
greedy algorithm that incorporates objects' depth and velocity in addition to
their 2D displacement for object association, resulting in more accurate object tracking.

Our experiments on the challenging nuScenes dataset \cite{caesar2020nuscenes} show 
that \method{} outperforms all other image-based tracking methods on the nuScenes
benchmark, as well as the nuScenes' baseline LiDAR-based method AB3DMOT \cite{weng2019baseline}. 
Our fusion-based object tracking algorithm achieves 20.0\% AMOTA,
outperforming CenterTrack \cite{zhu2018online} by a factor of 4, while running at 
28 frames per second.

\section{Related Work}
Object tracking methods have many applications in different computer vision tasks
such as autonomous driving, surveillance and activity recognition. Most 
existing methods on \ac{mot} use the tracking-by-detection approach
\cite{zhu2018online,xu2019spatial,fang2018recurrent}, relying on the performance
of an underlying object detection algorithm
\cite{ren2017accurate,ren2015faster,yang2016exploit} 
and focusing on improving the association between detections. One major drawback
in this approach is that the association task does not utilize the valuable features
extracted in the detection step. More recently, the \textit{joint detection and tracking} 
approach is trending for \ac{mot} where an existing object detection network is
converted into an object tracker to accomplish both tasks in the same framework
\cite{feichtenhofer2017detect,bergmann2019tracking,kang2017object}. Our method 
belongs to this category.

From another perspective, \ac{mot} algorithms can be split into batch and online
methods. Batch methods use the entire sequence of frames to find the global optimal
association between the detections. Most methods in this category are based on 
optical flow algorithms and create a flow graph from the entire sequence \cite{schulter2017deep,zhang2008global}. Online methods, on the other hand,
only use the information up to the current frame for tracking objects. Many of 
these algorithms generate a bipartite graph matching problem which is solved 
using the Hungarian algorithm \cite{bewley2016simple}. More modern methods in 
this category use deep neural networks to solve the association problem 
\cite{baser2019fantrack,weng2020gnn3dmot}. Our proposed algorithm is an online 
method and does not require any knowledge from future frames.

\ac{mot} methods can also be divided into 2D and 3D 
categories. Most 3D \ac{mot} methods are developed as an extension of existing 2D 
tracking models, with the distinction that input detections are in the 3D space 
rather than the 2D image plane. Some of the 3D \ac{mot} methods use LiDAR point 
clouds \cite{weng20203d} or a combination of point clouds and images 
\cite{zhang2019robust} as their inputs.

\subsection{2D Multi-Object Tracking}
DeepSORT \cite{wojke2017simple} uses an overlap-based association method with a 
bipartite matching algorithm, in addition to appearance features extracted by a 
deep network. In \cite{feichtenhofer2017detect} authors use the current and previous
frames as inputs to a siamese network that predicts the offset between the bounding 
boxes in different frames. Tracktor \cite{bergmann2019tracking} exploits the 
bounding box regression of the object detector network to directly propagate the 
region proposals' identities, which eliminates the need for a separate association 
step. Since it is assumed that the bounding boxes have a large overlap between 
consecutive frames, low frame-rate sequences would require a motion model in this 
approach. Zhu et al. \cite{zhu2017flow} propose flow-guided feature aggregation, where
an optical flow network estimates the motion between the current and previous frames. The 
feature maps from previous frames are then warped to the current frame using the 
flow motion and an adaptive weighting network is used to aggregate and feed them 
into the detection network. Integrated detection \cite{zhang2018integrated} 
proposes an early integration of the detection and tracking tasks, where the outputs
of the object detector are conditioned on the tracklets computed over the prior 
frames. A bipartite-matching association method is then used to associate the 
bounding boxes. 

\subsection{3D Multi-Object Tracking}
Hu et al. \cite{hu2019joint} combine 2D image-based feature association
and 3D LSTM-based motion estimation for 3D object tracking. Their method leverages
3D box depth-ordering matching and 3D trajectory prediction to improve instance 
association and re-identification of occluded objects.
Weng et al. \cite{weng20203d} propose a real-time \ac{mot} system called AB3DMOT
that uses LiDAR point clouds for object detection and a combination of Kalman filter
and the Hungarian algorithm for state estimation and data association.
CenterTrack \cite{zhou2020tracking}
takes a pair of images and detections from prior frames as input to a point-based 
framework, where each object is represented by the center point of its bounding 
box. The network estimates an offset vector from the center point of objects in 
the current frame to their corresponding center points in the previous frame, 
and uses a greedy algorithm for object association. 
Besides images and point clouds, some methods use map information to improve 
the tracking performance for autonomous driving applications. Argoverse
\cite{chang2019argoverse} uses detailed map information such as lane direction, 
ground height and drivable area to improve the accuracy of 3D object tracking.

\begin{figure*}[ht!]
\includegraphics[width=0.99\textwidth]{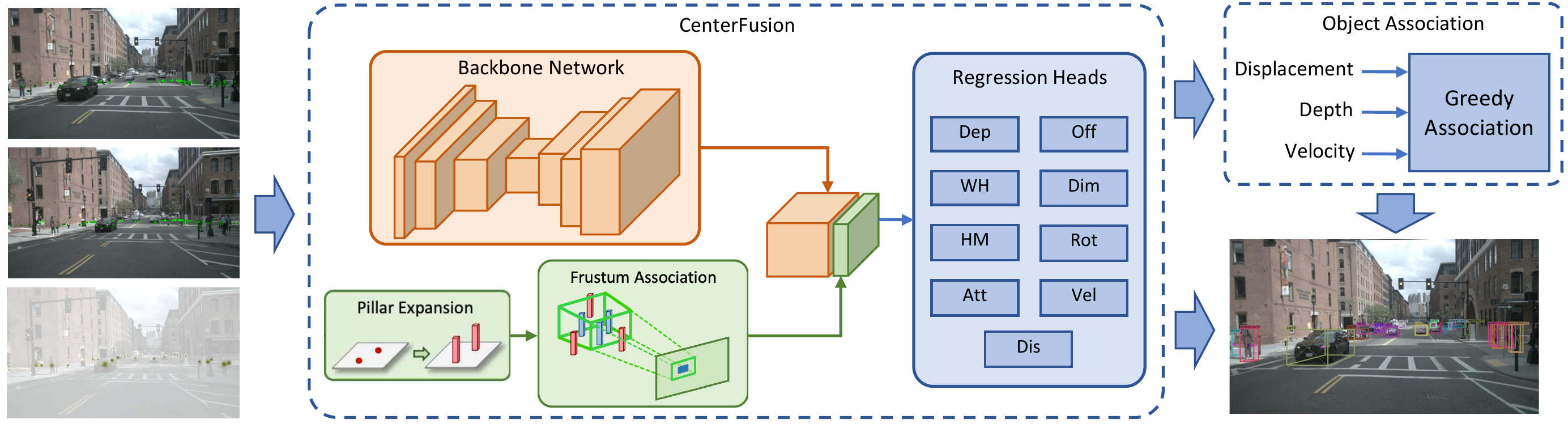}
\caption{\method{} network architecture. The inputs to the network are shown on the 
    left which includes the current image and radar point clouds (top), the previous 
    image frame and radar point clouds (middle) and the previous detection results in
    the form of class-agnostic heatmaps (bottom). The radar point clouds are shown on
    the input images. An additional regression head (``Dis'') is added to the model, 
    which uses the fused radar and image features to predict
    objects displacement in consecutive frames. The greedy algorithm in the association step
    uses the displacement, depth and velocity of each object to associate it to previous
    detections. The output is 3D bounding boxes and track IDs for all detected objects.}

\label{fig:architecture}
\end{figure*}
\section{Preliminaries}
Our proposed 3D tracking algorithm is based on the CenterFusion 3D object detection
algorithm \cite{nabati2020centerfusion}. 
CenterFusion takes an image
$I\in \mathbb{R} ^{W\times H \times 3}$ and a set of radar detections 
$P_i = (x^i, y^i, z^i, v_x^i, v_y^i)$ where $(x^i, y^i, z^i)$
are the coordinates of the point $i$ in the radar point cloud, and 
$(v_x^i, v_y^i)$ are the radial velocities in the $x$ and $y$ directions, respectively.
These coordinates are according to the vehicle coordinate system, where $x$ is forward,
$y$ is to the left and $z$ is upward from the drivers point of view.

CenterFusion first uses a center point detection method called CenterNet \cite{zhou2019objects}
to detect the centerpoint
of objects by estimating a heatmap 
$\hat{Y} \in [0,1]^{\frac{W}{R} \times \frac{H}{R} \times C}$
where $R$ is the down-sampling factor and $C$ is the number of object categories in 
the dataset. The local maxima in the estimated heatmap $\hat{Y}$ correspond
to the centers of detected objects in the image. The ground truth heatmap $Y$ is 
generated by rendering a Gaussian-shaped peak at the center points of each
object, calculated as the center point of their corresponding bounding box.

The network uses regression layers to generate preliminary 3D 
bounding boxes for all objects, then associates the radar detections to these preliminary 
3D detections using a frustum-based association method. To do this, the radar detections
are first expanded into pillars with predefined dimensions. A frustum is then formed 
around each detected object, and radar pillars inside the frustum are associated with
that object. If there are multiple radar pillars inside the frustum, the closest one
is kept and others are discarded.

Based on the association results, the depth
and velocity of the radar detection are mapped to their corresponding objects on the image.
These values are represented as separate heatmap channels and are concatenated 
to the image-based features. These fused features are then
used to improve the preliminary detection results by re-calculating the depth, size,
orientation and other object attributes. Additionally, a velocity vector is estimated 
for every detected object.

CenterFusion uses an objective function based on the focal loss, defined as:

\small
\begin{equation*}
    L_{i} = \frac{1}{N} \sum_{xyc}
    \begin{cases}
        (1 - \hat{Y}_{xyc})^{\alpha} 
        \log(\hat{Y}_{xyc}) & \! Y_{xyc}=1\vspace{2mm}\\
        (1-Y_{xyc})^{\beta} 
        (\hat{Y}_{xyc})^{\alpha}\log(1-\hat{Y}_{xyc})
        & \!\text{otherwise}
    \end{cases},
\end{equation*}
\normalsize
where $N$ is the number of objects, 
$Y \in [0,1]^{\frac{W}{R} \times \frac{H}{R} \times C}$ is the annotated 
objects' ground-truth heatmap and $\alpha$ and $\beta$ are the hyper-parameters
of the focal loss. After detecting objects' center point, different regression 
heads are used to regress to size, orientation, depth and velocity of the detected 
objects.

\section{\method}

We follow CenterTrack \cite{zhou2020tracking} and approach the tracking problem
from a local perspective, where an object's identity is preserved across consecutive
frames without re-establishing associations if the object leaves the frame. We use 
both camera and radar data from the previous frame to improve the ability to track 
occluded objects in the current frame. \method{} uses the fused radar and image
features to estimate objects' displacement in consecutive frames, which is used for 
object association through time. In the association step, a greedy algorithm 
is proposed that leverages objects' velocity and depth information in addition to their 
2D displacement for accurate association through time.

The next section presents our problem formulation. Section 
\ref{sec:detNet} describes the underlying detection network CenterFusion, 
which is modified to use the previous frame as an additional input and also 
estimate objects' displacements between consecutive frames. Finally, Section
\ref{sec:objAssoc} discusses the greedy algorithm used to associate detected 
objects.

\subsection{Problem Formulation} \label{sec:probForm}
The inputs to \method{} are the current and previous image frames 
$I^{(t-1)}, I^{(t)} \in \mathbb{R}^{W \times H \times 3}$, the current and previous
radar detections $P^{(t-1)}, P^{(t)} \in \mathbb{R}^{N \times 5}$ where $N$ 
is the number of radar detections, and the tracked objects from the previous frame
$T^{(t-1)}=\{b_0^{(t-1)}, b_1^{(t-1)}, ...\}$. The tracked objects are represented
by $b=(p, d, v, w, id)$ where $p \in \mathbb{R}^2$ is the object's center location,
$d \in \mathbb{R}$ is the object's depth, 
$v \in \mathbb{R}^2$ is object's velocity, $w \in [0, 1]$ is the detection confidence
and $id$ is an integer representing the unique identity of the tracked object.
For every frame, the goal is to detect and track objects 
$T^{(t)}=\{b_0^{(t)}, b_1^{(t)}, ...\}$ and assign a consistent $id$ to the objects
in consecutive frames. The detection and association of objects are done in a single
deep network trained end-to-end.

\subsection{Detection Network} \label{sec:detNet}
The overall network architecture is shown in Fig. \ref{fig:architecture}.
The CenterFusion network is modified to take as input the current image frame 
$I^{(t)}$ and radar 
detections $P^{(t)}$, in addition to the previous image frame $I^{(t-1)}$, 
radar detections $P^{(t-1)}$ and
detected objects. The outputs are 3D bounding boxes for all detected objects and 
an absolute velocity for each object, reported in the $x$ and
$y$ directions in the vehicle's
coordinate system. The previous detections are represented as a single channel heatmap 
using a 2D Gaussian kernel.
Including the previous image, radar detections and detected objects helps the 
network to better estimate the location of objects in the current frame. The
radar information from previous frame further improves the ability of network 
to detect objects even if the visual evidence is not present due to occlusion.

Besides the object detection results for the current frame, the modified network
also estimates the 2D displacement of the detected objects between the current
and previous frames, using the concatenated radar and image features. Having the
radar depth and velocity information in addition to the image features from the 
current and previous frames helps the network to generate more accurate object
displacement predictions.

\subsection{Object Association} \label{sec:objAssoc}
\begin{figure}[t]
\includegraphics[width=0.99\linewidth]{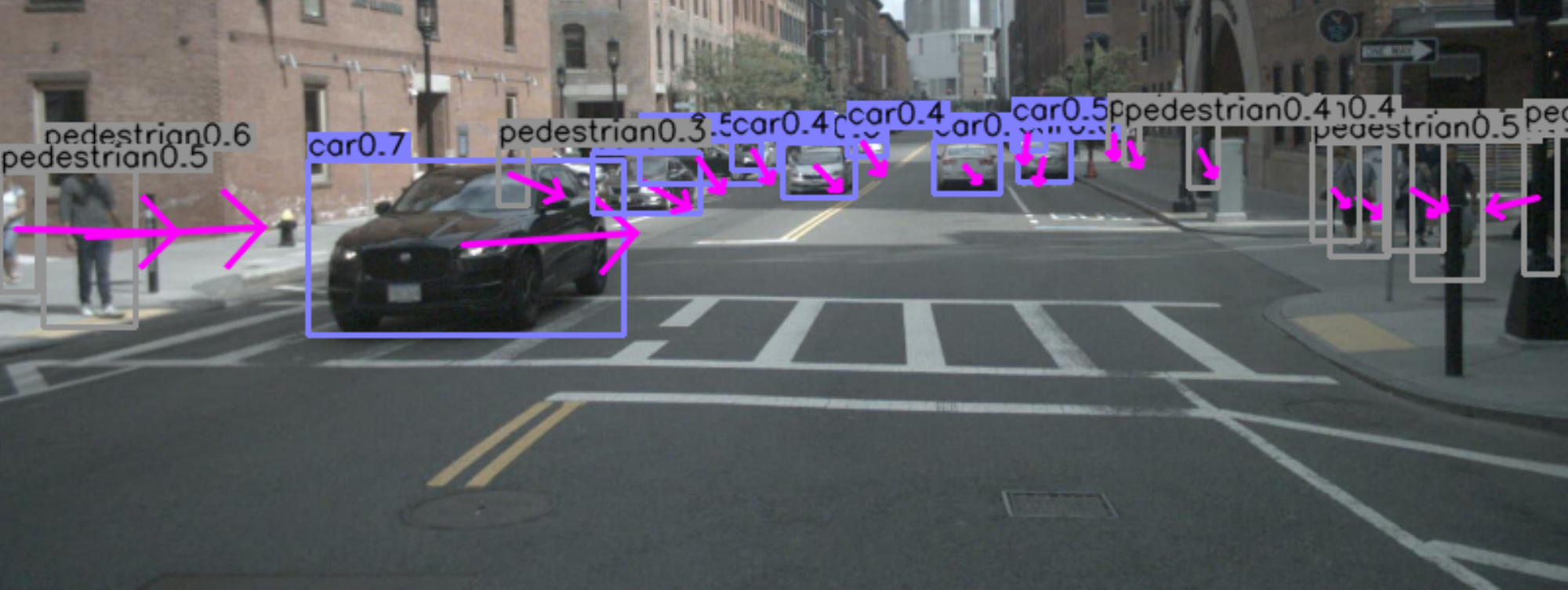} \\[6pt]
\includegraphics[width=0.99\linewidth]{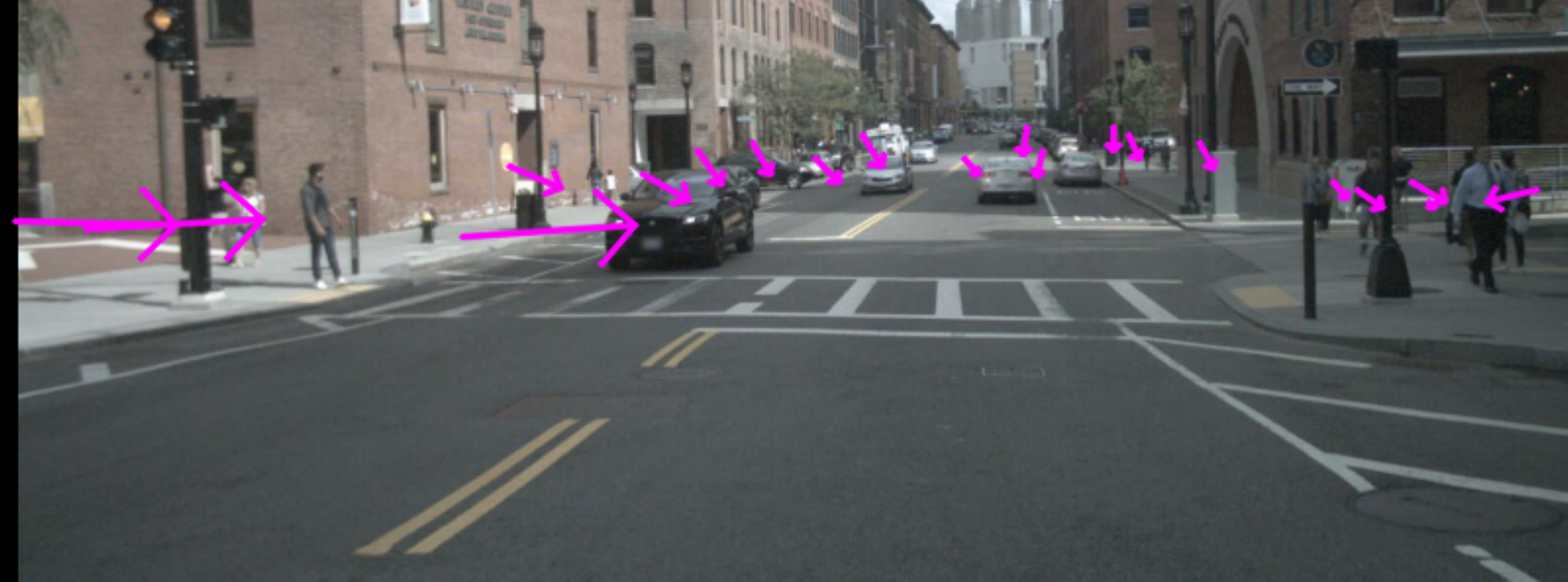}
\caption{Top: Object displacement from the previous frame, represented by arrows pointing 
    from the center of each object in the current frame to the estimated center of the 
    same object in the previous frame. 
    Bottom: Previous frame. Displacement arrows re-drawn for comparison.}
\label{fig:displacement}
\end{figure}

We use a greedy algorithm to associate the detected objects over time. The detected 
objects are represented by $a=(p,d,v,c)$ where 
$p \in \mathbb{Z}^2$ is the object's center in pixels, $d \in \mathbb{R}$ is the 
object's depth, $v \in \mathbb{R}^2$ is the object's velocity, and 
$c \in C$ is the object's category.
Similar to \cite{zhou2020tracking}, the displacement is
calculated by a regression layer in the form of two output channels 
$\hat{D}^{(t)} \in \mathbb{R}^{\frac{W}{R} \times \frac{H}{R} \times 2}$ 
representing the displacement of the center of the objects on the image, as shown
in Fig. \ref{fig:displacement}. Similar to the other regression heads, the L1 loss
is used as the objective function to train this layer.

To associate objects across time, we define a cost function based on the objects' depth,
velocity and displacement on the image. 
This cost function is defined as: 

\small
\begin{equation*}
    Cost_{t,t-1} = 
    \begin{cases}
        \alpha \cdot \mathcal{L}_{pixel} + \beta \cdot \mathcal{L}_{depth} + \delta \cdot \mathcal{L}_{velocity} & \! c_t = c_{t-1} \\
        \infty                                    & \! c_{t} \neq c_{t-1}
    \end{cases}
\end{equation*}

\begin{align*}
    \mathcal{L}_{pixel} &= (x_t-x_{t-1})^2 + (y_t-y_{t-1})^2\\
    \mathcal{L}_{depth} &= \left(d_t - d_{t-1}\right)^2\\
    \mathcal{L}_{velocity} &= (vx_t-vx_{t-1})^2 + (vy_t-vy_{t-1})^2\\
\end{align*}
\normalsize
where $x$, $y$ is the object's center, $d$ is the object's depth, and $vx$, $vy$ are the
velocity of the object in $x$ and $y$ directions respectively. 
$\alpha,\beta,\delta \in \mathbb{R}^+$, are tunable parameters.

For every detected object at position $p$, we look for prior detections within a 
radius $r$ from $p - D_p$. If there are unmatched prior detections at that position,
we calculate the above cost function to determine the distance between these detections,
and match the object with the previous detections with the lowest cost. For every 
unmatched detection, a new track is created.

\begin{table*}[t]
    \centering
    \caption{Evaluation on the nuScenes test set. We compare to published works 
    on the nuScenes benchmark, including vision-based and LiDAR-based methods. AMOTA is the main metric used in the nuScenes benchmark. Other metrics are defined in \cite{caesar2020nuscenes}.}
    \begin{tabular}{@{\extracolsep{6pt}}l
        c@{\hskip0.0cm}
        c@{\hskip0.0cm}
        c@{\hskip0.0cm}
        c@{\hskip0.1cm}
        >{\columncolor[gray]{0.8}}c@{\hskip0.1cm}
        c@{\hskip0.0cm}
        c@{\hskip0.0cm}
        c@{\hskip0.0cm}
        c@{\hskip0.0cm}
        c@{\hskip0.0cm}}
        \hline
        & \multicolumn{3}{c}{Modality} & & & & & & & \\\cline{2-4}
        Method & Cam & Rad & LiDAR & Time(ms) & AMOTA & AMOTP & MOTAR & MOTA & MOTP & Recall\\ 
        \hline
        AB3DMOT \cite{weng2019baseline} + Mapillary &  &  & \checkmark & - & 0.018 & \textbf{1.790} & 0.091 & 0.020 & \textbf{0.903} & 0.353 \\
        AB3DMOT \cite{weng2019baseline} + PointPillars &  &  & \checkmark & - & 0.029 & 1.703 & 0.243 & 0.045 & 0.824 & 0.297\\
        AB3DMOT \cite{weng2019baseline} + Megvii &  &  & \checkmark & - & 0.151 & 1.501 & \textbf{0.552} & \textbf{0.154} & 0.402 & 0.276 \\
        CenterTrack \cite{zhou2020tracking} & \checkmark &  &  & 45 & 0.046 & 1.543 & 0.231 & 0.043 & 0.753 & 0.233 \\
        CenterTrack \cite{zhou2020tracking} + Megvii & \checkmark &  & \checkmark & 45 & 0.108 & 0.989 & 0.267 & 0.085 & 0.349 & 0.412 \\
        \hline
        Ours & \checkmark & \checkmark &  & \textbf{35} & \textbf{0.200} & 1.292 & 0.353 & 0.151 & 0.766 & \textbf{0.420} \\
        \hline
    \end{tabular}
    \label{table:scores}
\end{table*}

\begin{table*}[t]
    \centering
    \caption{Per-class AMOTA results.}
    \begin{tabular}{@{\extracolsep{6pt}}l@{\hskip0.2cm}
        c@{\hskip0.0cm}
        c@{\hskip0.1cm}
        c@{\hskip0.1cm}
        c@{\hskip0.1cm}
        c@{\hskip0.1cm}
        c@{\hskip0.1cm}
        c@{\hskip0.1cm}
        c@{\hskip0.0cm}
        c@{\hskip0.0cm}
        c@{\hskip0.0cm}
        c@{\hskip0.1cm}}
        \hline
        & \multicolumn{3}{c}{Modality} & \multicolumn{7}{c}{AMOTA} \\
        \cline{2-4} \cline{5-11}
        Method & Cam & Rad & LiDAR & Car & Truck & Bus & Trailer & Pedest. & Motor. & Bicycle \\
        \hline
        AB3DMOT \cite{weng2019baseline} + Mapillary & & & \checkmark & 0.125 & 0.000 & 0.000 & 0.000 & 0.000 & 0.000 & 0.000 \\
        AB3DMOT \cite{weng2019baseline} + PointPillars & & & \checkmark & 0.094 & 0.000 & 0.066 & 0.000 & 0.039 & 0.000 & 0.000 \\
        AB3DMOT \cite{weng2019baseline} + Megvii & & & \checkmark & 0.278 & \textbf{0.013} & \textbf{0.408} & \textbf{0.136} & 0.141 & 0.081 & 0.000 \\
        CenterTrack \cite{zhou2020tracking} & \checkmark & &  & 0.202 & 0.004 & 0.072 & 0.000 & 0.030 & 0.011 & 0.000 \\
        CenterTrack \cite{zhou2020tracking} + Megvii & \checkmark & & \checkmark  & 0.341 & 0.012 & 0.256 & 0.000 & 0.142 & 0.005 & 0.000 \\
        \hline
        Ours & \checkmark & \checkmark &   & \textbf{0.546} & 0.000 & 0.107 & 0.075 & \textbf{0.346} & \textbf{0.206} & \textbf{0.114} \\
        \hline
    \end{tabular}
    \label{table:classBased}
\end{table*}

\section{Experiments}
\subsection{Dataset and Evaluation Metrics}
We evaluate our method on the nuScenes dataset \cite{caesar2020nuscenes}, a 
large-scale dataset for autonomous driving with annotations for 3D object
detection and tracking containing camera, radar and LiDAR data. It provides 1000
different sequences from which 700 sequences are used for training, 150 sequences
for validation and 150 sequences for testing. Each sequence is comprised of 40
annotated frames, each containing camera, radar and LiDAR samples. Samples
are obtained from 6 different cameras and 5 different radars.

The main evaluation metric used in the nuScenes benchmark, AMOTA, is a weighted average
of the recall-normalized \ac{mota} metric at different recall thresholds:

\small
\begin{align*}
    &MOTAR = \max (0, 1-\frac{IDS_r + FP_r + FN_r - (1-r)*P}{r*P})\\[4pt]
    &AMOTA = \frac{1}{n-1} \sum_{r\in \{\tfrac{1}{n-1}, \tfrac{2}{n-1}, ..., 1\}} MOTAR
\end{align*}
\normalsize
where $r$ is the recall threshold, $IDS_r$ is the number of identity switches,
$FP_r$ is the number of false positives, $FN_r$ is the number of false negatives
and $P$ is the total number of annotated objects in all frames. 

\subsection{Implementation Details}
Following CenterFusion \cite{nabati2020centerfusion}, we use an input resolution
of $800 \times 448$ and apply horizontal flipping
and random shifts for regularization. The \ac{dla} network is used as the backbone
for extracting image features, optimized with the Adam \cite{kingma2014adam}. 
The \method ~network is trained for 60 epochs with a 
batch size of 24 and a learning rate of 1.2$e$-4, starting from a pre-trained
CenterFusion network trained for 170 epochs. We train on a machine with an Intel
Xeon E5-1650 CPU and two Quadro P5000 GPUs. The reported tracking runtime 
is obtained on a machine 
with an Intel Xeon E5-1607 CPU and a TITAN X GPU. We re-implement the Frustum 
Fusion algorithm in CenterFusion to run in parallel and improve the 
overall runtime of the detection network.

\section{Results}
Table \ref{table:scores} compares our method with some of the other published 
methods in the nuScenes object tracking benchmark. Specifically, we compare our method
with the CenterTrack \cite{zhou2020tracking} algorithm using both image-based and LiDAR 
based detection results, as well as the AB3DMOT \cite{weng2019baseline} with three 
different LiDAR-based object detection algorithms. According to the table, \method{} 
outperforms both methods by achieving an AMOTA score of 20.0$\%$, improving the 
vision-based CenterTrack algorithm by about 15$\%$ (by a factor of 4) and the LiDAR-based 
CenterTrack algorithm by 9.2$\%$. \method{} also outperforms CenterTrack in the MOTAR, 
MOTA, MOTP and Recall metrics. 

Given the similarity of the tracking algorithms in 
\method{} and CenterTrack, these results demonstrate the effect of utilizing radar
data in both detection and tracking stages. Both methods use a greedy algorithm for 
associating objects, but \method{} also takes advantage of the depth and velocity 
of the detected objects to better associate them through time. Additionally, the velocity
data provided by the radar enables the network to predict the objects' displacement 
in the image more accurately, further improving object association.

Table \ref{table:classBased} shows AMOTA for each class. According to the results, \method{}
significantly outperforms all the other methods in the Car, Pedestrian, Motorcycle and Bicycle 
categories, while AB3DMOT with the Megvii detector performs better in the Truck, Bus and 
Trailer categories. Note that all categories where \method{} is outperformed are large
objects, namely Truck, Bus and Trailer. One explanation could be the fact that it is more
difficult for the underlying fusion algorithm to correctly associate many radar detections
obtained from these large objects to their corresponding 3D bounding boxes, resulting in 
lower accuracy in estimated depth and velocity for these objects. 

On average, our method achieves a runtime of 35ms per image (28 fps), which makes 
it suitable for the real-time autonomous driving applications.

\section{Conclusions}
In this work, we presented an online and end-to-end object detection and tracking
method based on radar and camera sensor fusion. The proposed method uses the fused
radar and image features to detect objects and also estimate their displacement 
from the previous frame. To associate objects across time, we proposed a greedy 
algorithm that uses a cost function incorporating objects' depth, velocity and 
displacement, which makes our tracking algorithm very robust to overlapping and 
occluded objects. Our proposed method is currently the only tracking algorithm 
based on radar and image information on the nuScenes tracking benchmark, outperforming
all published vision-only methods as well as the nuScenes' baseline LiDAR-based method.

\addtolength{\textheight}{-6.5cm}   %

\section{Acknowledgement}
This work was presented at the 3D-Deep Learning for Autonomous Driving Workshop (WS15), IV2021.

\bibliographystyle{IEEEtran}
\bibliography{references}

\end{document}